\documentclass[11pt,a4paper]{article}
\usepackage[hyperref]{naaclhlt2019}
\usepackage{times}
\usepackage{latexsym}
\usepackage{amssymb}
\usepackage{amsmath}
\usepackage{graphicx}
\usepackage{caption}
\usepackage{subcaption}
\usepackage{comment}
\usepackage{url}

\aclfinalcopy 

\title{Adding Interpretable Attention to Neural Translation Models Improves Word Alignment}

\author{Thomas Zenkel, Joern Wuebker, John DeNero \\
   Lilt, Inc. \\
  {\tt first\_name@lilt.com}
}

\date{}

\begin{document}
\maketitle

\begin{abstract}
Multi-layer models with multiple attention heads per layer provide superior translation quality compared to simpler and shallower models, but determining what source context is most relevant to each target word is more challenging as a result.
Therefore, deriving high-accuracy word alignments from the activations of a state-of-the-art neural machine translation model is an open challenge.
We propose a simple model extension to the Transformer architecture that makes use of its hidden representations and is restricted to attend solely on encoder information to predict the next word.
It can be trained on bilingual data without word-alignment information.
We further introduce a novel alignment inference procedure which applies stochastic gradient descent to
directly optimize the attention activations towards a given target word.
The resulting alignments dramatically outperform the na\"ive approach to interpreting Transformer attention activations, and are comparable to Giza++ on two publicly available data sets.

\end{abstract}

\section{Introduction}

Self-attention-based sequence-to-sequence models \cite{vaswani2017attention} have recently emerged as the state of the art in neural machine translation. The decoder of the network typically consists of multiple layers, each with several attention heads. 
This makes it hard to interpret the attention activations and extract meaningful word-alignments. 
As a result, the most widely used tools to obtain word alignments are still based on the statistical IBM word alignment models which were introduced more than two decades ago. 
In this work we describe a simple modeling extension as well as a novel inference procedure that are capable of extracting alignments with a quality comparable to Giza++ from the self-attentional Transformer architecture.

Word alignments extracted as a by-product of machine translation have a number of applications.
For example, they can be used to inject an external lexicon into the inference process to improve
the translations of low-frequency content words \cite{Arthur_2016}.
Another use of word alignments is to project annotations from the source sentence to the target sentence. For example, if part of the source sentence is highlighted in the source document, the corresponding part of the target should be highlighted as well. In localization, all formatting and annotation is stored as \emph{tags} over spans of the source, and word alignments can serve to place those tags automatically into the target.

The most widely used tools for word alignment are Giza++ \cite{Och_2003}, which uses the IBM Model 4 in its default
setting, and FastAlign \cite{dyer2013simple}, which relies on a re-parameterized version of the IBM Model 2. Previous work on neural models for word alignment either require complicated training approaches
\cite{Tamura_2014} or rely on supervised training, the training data for which is generated with
one of the tools mentioned above \cite{alkhouli2018alignment,Peter_2017}.

This work extends the Transformer architecture with a separate alignment layer on top of the decoder sub-network.
It does not contain skip connections around the encoder-attention module, so that it is trained to
predict the next target word based on a linear combination of the encoder information. This encourages
the alignment layer to focus its attention activations on relevant source words for a given target word.
As a result we have two separate output layers, each of which defines a probability distribution over the
next target word. During inference the add-on output layer is ignored and only the attention activations are
computed, which are interpreted as a distribution over word alignments.

However, the attention mechanism still ignores all future target information, including the target token
for which it computes the attention activations. In order to query the model 
w.r.t. the aligned target word to further improve the resulting alignments, we directly optimize the attention activations to
maximize the likelihood of the target word using stochastic gradient descent (SGD).

Our approach has a number of desirable properties:
\begin{itemize}
    \item The model can be trained on the same data as the translation model in an unsupervised fashion.
    \item Both the alignment and the translation model are incorporated into a single network.
    \item Training is done by fine-tuning an existing translation model, significantly reducing overall training cost.
    \item The extension is straightforward and easy to implement.
\end{itemize}

We validate our approach on three hand-aligned, publicly available data sets and compare the alignment error rate (AER) to a na\"ive baseline, FastAlign and Giza++. On the French-English task our method
improves the alignment quality by a factor of three over the na\"ive baseline. With the application
of the grow-diagonal heuristic for bidirectional alignment merging \cite{koehn2005edinburgh}, we achieve results that are
comparable to Giza++ on two of the three data sets.

\section{Machine Translation Model}
All neural machine translation (NMT) models in this work are based on the Transformer architecture
introduced by \citet{vaswani2017attention} . It follows the encoder-decoder paradigm \citep{sutskever2014sequence} and is composed of two sub-networks.
The encoder network transforms the source sentence into a high-dimensional continuous space 
representation of the sentence. The decoder network uses the output of the encoder to compute a probability
distribution over target language sentences. Different from the previously most widely used
recurrent networks, it relies entirely on the attention mechanism to incorporate context.
The attention function \citep{bahdanau2014neural} is a major factor in the recent success of NMT and
provides a mechanism to create a fixed-length context vector as a weighted sum of a variable number of input vectors.

In the Transformer architecture, each layer of the encoder network consists of two sub-layers,
namely a self-attention module and a feed-forward module. Decoder layers are made up of a self-attention
module, an encoder-attention module and a feed-forward module. In the decoder the self-attention 
sub-network is masked so that only left-hand context is incorporated. In contrast to \citet{vaswani2017attention}, we mask the attention to not attend to the end-of-sentence token. The encoder-attention always
uses the output of the final encoder layer as input.

The attention function applied in the Transformer is a multi-head scaled dot-product variant.
Given a query $Q$, a set of $k$ key-value pairs $K, V$ with $Q \in \mathbb{R}^{d}$ and $V, K \in \mathbb{R}^{k \times d}$, we calculate a single attention vector $A \in \mathbb{R}^{k}$ as follows:
\begin{equation}
    \mathsf{calcAtt}(Q, K) = \mathsf{softmax}(\frac{Q \cdot K^{T}}{\sqrt{d}})
\end{equation}

With the resulting attention activations $A$ we calculate the weighted sum of the values:
\begin{equation}
    \mathsf{applyAtt}(A, V) = A \cdot V
\end{equation}

Multi-head attention is computed from the concatenation of $m$ context vectors $h_i$ calculated with lower-dimensional attention
functions and the parameter matrices $W^O \in \mathbb{R}^{d \times d}$, $W_i^Q, W_i^K, W_i^V \in \mathbb{R}^{d \times \frac{d}{m}}$:
\begin{align}
   \mathsf{mHead}(Q,K,V) &= \mathsf{concat}(h_i)W^O \label{eqn:mhead} \nonumber\\
   h_i(A_i,V) &= \mathsf{applyAtt}(A_i, V W_i^V) \\
   A_i(Q,K) &= \mathsf{calcAtt}(Q W_i^Q, K W_i^K) \nonumber
\end{align}
That is for each head the query and the set of key-value pairs get linearly projected before getting fed to the individual attention heads. The resulting output of the individual heads gets concatenated, linearly projected and fed into the next layer.

We use an encoder with 6 and a decoder with 3 layers. The decoder sublayers are simplified versions of those described by \citet{vaswani2017attention}: The filter sub-layers perform only a single linear transformation, and layer normalization is only applied once per decoder layer after the filter sub-layer.

\subsection{Average Transformer Attentions}
The straightforward method to extract alignments from the Transformer architecture is to use its attention activations. As our baseline we average all attention matrices $A_i$ into the matrix $A \in \mathbb{R}^{s \times t}$ to extract a soft-attention between subword units. $s$ denotes the source length and $t$ the target length. To extract hard alignments between subword units, we select the source unit with the maximal attention value for each target unit. We align a source word to a target word if an alignment between any of its subwords exists.
We argue that this procedure is not the best approach, as it does not encourage the soft attentions to correspond to useful alignments. 

\section{Related Work}

\subsection{Statistical Models}
The most commonly used statistical alignment models directly build on the lexical translation models of \citet{brown1993mathematics}, which are also referred to as IBM models. 
A popular tool is FastAlign \citep{dyer2013simple}, a reparameterization of IBM Model 2, which is known for its usability and speed. Giza++ \citep{Och_2003}, which is based on IBM Model 4, provides a solid benchmark in terms of AER results. In our experiments we run the MGIZA++ \cite{gao2008parallel}, a parallel implementation of Giza++, and FastAlign toolkits with default parameters as a baseline.

The IBM models are composed of an alignment and a lexical model component. The alignment
component is unlexicalized. The lexical component models the likelihood for each source word
based on a single target word, i.e. it is conditionally independent of the source and target context.
We argue that this assumption is a disadvantage over neural approaches, which commonly encode the content and the context of each word in a continuous representation. In this work we make use of both the source and the target context to infer word alignments.

\subsection{Neural Models}
The neural approaches that affect the attention of the model and therefore influence the alignments can be categorized into two groups depending on their goal: Improving translation quality of the machine translation system or solely focusing on the generation of alignments.

\citet{Nguyen_2018} add a linear combination of source embeddings to the decoder output to improve the prediction of the next target word in an attention-based translation model. This encourages the model to attend to a useful source word and avoids that the resulting alignments are shifted by one word compared to human judgement.
\citet{alkhouli2018alignment} train a single alignment head of the Transformer with supervised data generated with Giza++, so that its attention directly corresponds to alignments. This improved attention is then used during inference to separate the translation objective into a lexical and an alignment component and improve dictionary-guided translations. \citet{Arthur_2016} add a lexical probability vector, a vector generated based on the information of a discrete lexicon, and use the attention vector to decide which source word's lexical probabilities the model should focus on.

\citet{Tamura_2014} directly predict alignments based on a recurrent neural network which is conditioned on both the source and the target sequences. By using noise contrastive estimation and tying weights of a forward and a backward model during training they are able to train this network while only relying on IBM Model 1 to generate negative examples. \citet{Peter_2017} build on an attention-based neural network to extract alignments. They achieve their best results by bootstrapping the attention with Giza++ alignments and by using target foresight, a technique that uses the target word during training to improve its attention.

\section{Alignment Layer}
 \label{sec:alignmentLayer}
 In order to train our alignment component in an unsupervised fashion, i.e. without word-aligned training data, we want to design it with the following property: A source token should be aligned to a target token if we can predict the target token based on a continuous representation of the corresponding source token.
 
\begin{figure}
    \centering
    \includegraphics[scale=0.33]{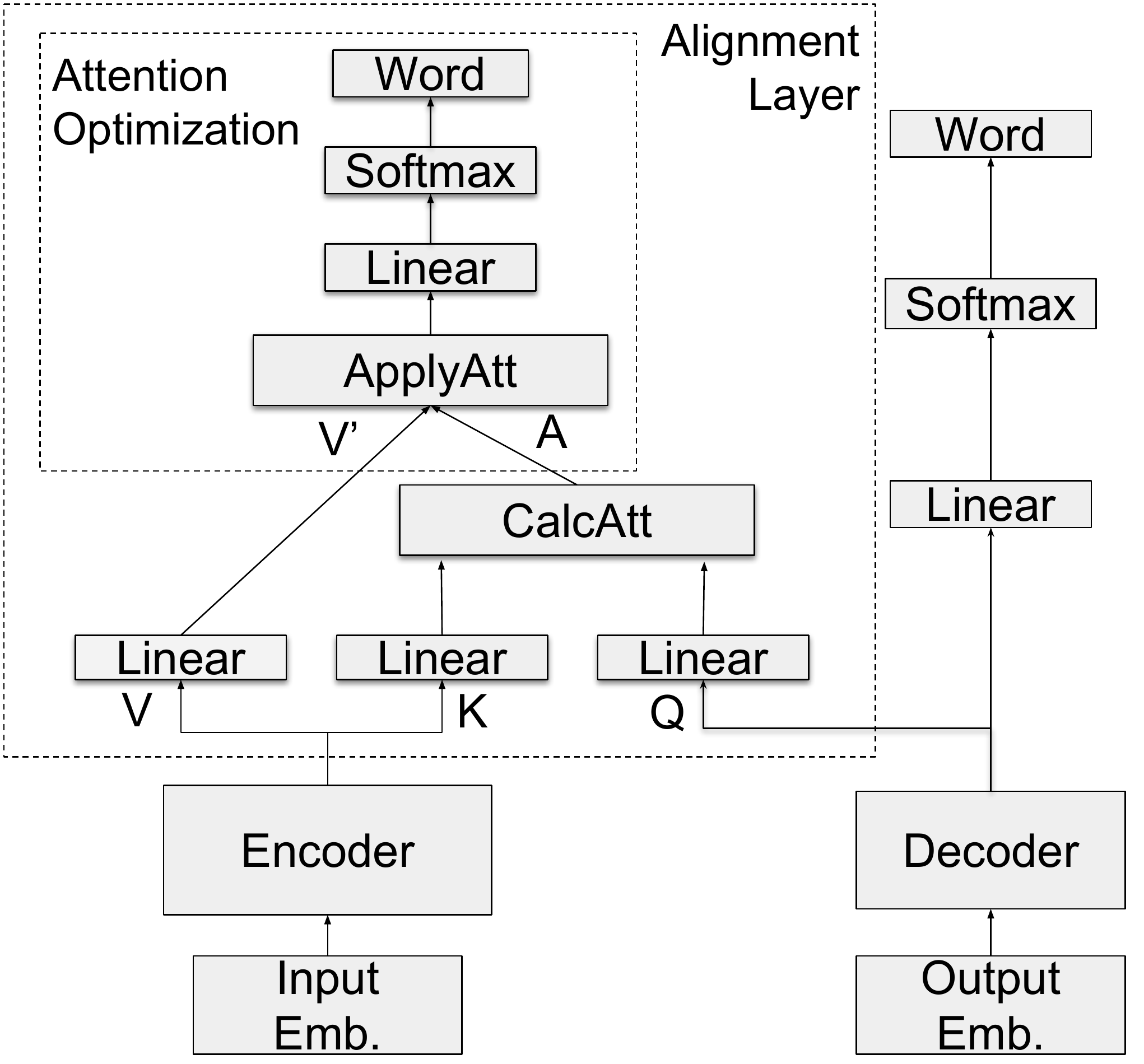}
    \caption{Architecture of the neural network. In addition to the Transformer we add an \emph{alignment layer}. During inference we optimize the attention weights $A$ of the sub-network \emph{attention optimization}.}
    \label{fig:training}
\end{figure}

 We achieve this by adding an alignment layer to the top of the decoder. As depicted in Figure \ref{fig:training} the complete model predicts the next target word twice: once with the original decoder, once based on the alignment layer.
 
 The alignment layer uses a single multi-head attention submodule as in Equation \ref{eqn:mhead} and is focusing its attention on the encoder. As the query $Q$ we use the decoder output, for the key-value pairs we use the same encoder input, i.e. $K=V$. 
 
 For now let us denote the vectors based on the hidden representation of the encoder, which we use as the keys and values, as $E$. We calculate the probability vector $p$ of the next word as follows:
 \begin{equation}
     p(Q, E) = \mathsf{softmax}(W h)
\end{equation}
with $W$ as the output projection matrix and $h$ being the output of the multi-head attention of Equation \ref{eqn:mhead}:
 \begin{equation}
     h(Q, E) = \mathsf{mHead}(Q, E, E)
 \end{equation}

 In contrast to a decoder layer of the Transformer, the alignment layer does not use a self-attention sub-layer and we \emph{do not apply any skip connections}. Thus the target word prediction of the alignment layer is forced to rely solely on the context vector $h$, a linear combination of the encoder-side representations.

Figure \ref{fig:training} summarizes the whole architecture. Note that the decoder output is masked, it only encodes the left-hand context.

For the encoder representation $E$ we want to encode both the content of the source words and the context in which they appear. Therefore, we experiment with the following options: Directly using the word embeddings, using the encoder output and using the average of the word embeddings and the decoder output as the encoder representation $E$. Table \ref{tab:alHeadInput} summarizes these options. 

\begin{table}[t!]
  \centering
  \begin{tabular}{ c c}
  Method & Explanation \\
  \hline
  \hline
  Avg & average over all attentions \\
  Word & word emb. \\
  Add & (word emb. + enc. out.) / 2.0 \\
  Enc & enc. out. \\
  \hline
  Add+SGD & initialize with forward path \\
  Rand+SGD & initialize randomly  \\
  \hline
\end{tabular}
\caption{Input used as key and value in the alignment layer in our experiments as a function of the word embeddings (word emb.) and the output of the encoder (enc. out.). The second part summarizes different initialization schemes when learning the attention weights with SGD. }
\label{tab:alHeadInput}
\end{table}

We train the alignment layer by fine-tuning a fully trained translation model, keeping the parameters of the underlying Transformer network fixed. For the alignment layer we use multi-head attention with a single head throughout this paper.\footnote{We verified that this leads to slightly better results than two attention heads, while four heads performed considerably worse.}

\section{Attention Optimization based on Target Word}
Using the attention activations of the forward pass means that the word alignment to the $i$-th target word $e_i$ does not depend on the identity of $e_i$. However, it can be argued that this word is the most relevant information needed to select the correct alignment.
When performing the task of word alignment given both source and target sentence we already now the target word $e_i$. If we produce the alignment as a by-product of translation inference, it is likely that the prediction of the alignment layer and the actual target sentence are different. 

We hypothesize that we can improve the alignment if we can find attention activations that lead to a correct prediction of the target sentence. Given attention activations $A_i$ and the linearly transformed values $V'_i = E W_i^V$, we can rephrase the equations of Section \ref{sec:alignmentLayer}: 
\begin{align}
   p(A, V') &= \mathsf{softmax}(W x) \nonumber \\
   x(A, V') &= \mathsf{concat}(h_i) W^O \label{eqn:mheadopt} \\
   h_i(A_i, V') &= \mathsf{applyAtt}(A_i, V'_i) \nonumber
\end{align}

Therefore, the probability distribution $p$ only depends on $V'$, which we extract from the forward pass, and the attention activations $A$. We can optimize $A$ while evaluating the \emph{attention optimization} sub-network of Figure \ref{fig:training} with its only input $V'$. We treat $A$ as a weight matrix for the remainder of this section and will refer to it as the attention weights.

The $i$-th entry of the probability vector $p_i$ denotes the probability of the target word $e_i$. Therefore, we can formulate our objective of maximizing the probability of the target word $e_i$ with respect to $A$:

\begin{equation}
    \operatorname{max}_A p_i(A, E)
\end{equation}

We optimize the attention weights $A$ for each word of the target sequence while keeping all other parameters of the alignment layer fixed. By applying gradient descent, we iteratively update the attention weights towards the goal of maximizing the probability of the correct target word. This can be done in parallel for each word of the target sentence using the cross entropy loss. 

During optimization we relax the constraint for $A$ to be a valid probability, i.e. to sum up to 1. While we experimented with using the softmax function during optimization, we found that only applying the rectified linear unit (RLU, $max(A_i, 0)$) to guarantee non-negative activations before passing it to the $\mathsf{applyAtt}$ function is easier to optimize. 

This optimization procedure is related to feature visualization applied in image recognition \cite{erhan2009visualizing, olah2017feature}, which maximizes the output of a neuron with respect to the input image.

\subsection{Initialization of Attention Weights}
\label{subsec:sgdInitialization}
The question remains how we initialize the attention weights $A$. A straightforward option is to initialize them randomly. While it is probably useful to initialize them with valid probability vectors, we used a uniform distribution between 0.0 and 1.0 as our first option.

However, intuitively it seems more reasonable to start with attention weights that correspond to a good alignment and that might be closer to a good local minimum. It is possible to convert alignments between subwords directly to attention weights and therefore using a hypothesis of an arbitrary alignment model\footnote{That can be done by setting weights that represent alignments between a source and a target word to 1.0 and all other weights to 0.0.}. However, we restrict our experiments to improve the attention weights of the forward pass. Therefore, we run a forward pass of the complete Transformer network, extract the attention weights of the alignment layer and start the optimization process with these weights. 
 
 While tuning the parameters on the validation set, we found that applying three gradient descent steps with a learning rate of 1 leads to surprisingly good results, both in terms of predicting the correct word and the quality of the resulting alignments.

\section{Experimental Setup}
Our goal for the evaluation is to compare statistical alignment methods, namely FastAlign and Giza++, with the neural approach introduced in this paper. We attempt to do a fair comparison by using the same training data for all methods and standardize the pre-processing. We use publicly available training and test data for the following language pairs:  German-English, Romanian-English and French-English. All approaches are evaluated for both forward and reverse direction as well as by combining these with the grow-diagonal heuristic \cite{koehn2005edinburgh}, i.e. without using the finalize step. All hyper-parameters are tuned on the German-English task. We open-source our preprocessing pipeline and the baseline experiments using FastAlign and MGIZA++\footnote{\url{https://github.com/lilt/alignment-scripts}}.

\subsection{Training Data}
The training data we use has between 0.4 and 1.9 million parallel sentences. This makes it possible to train both statistical and neural methods in a reasonable amount of time. For German-English, we use the Europarl v8 corpus. For Romanian-English and English-French we follow \citet{Mihalcea_2003}. As the only exception we additionally use Europarl data for Romanian-English to increase the training data from 49k to 0.4M parallel sentences.

We preprocess the training data with the tokenizer from the Moses toolkit\footnote{\url{https://github.com/moses-smt/mosesdecoder/blob/master/scripts/tokenizer/tokenizer.perl}} and consistently lowercase all training data. For all neural approaches we apply byte pair encoding \cite{Sennrich_2016} with 10k merges, see Table \ref{tab:preprop} for an example. For FastAlign and Giza++ we concatenate the training and test data. 
Table ~\ref{tab:trainingdata} summarizes the training data we use.

\begin{table*}[t]
  \centering
  \begin{tabular}{ c c}
  Method & Text \\
  \hline
  Source  & Wir glauben nicht , da{\ss} wir nur Rosinen herauspicken sollten . \\
  Source BPE & wir \_gauben \_nicht \_, \_da{\ss} \_wir \_nur \_ro s inen \_her au sp ick en \_sollten \_. \\
  Target & We do not believe that we should cherry-pick .\\
  Target BPE & we \_do \_not \_believe \_that \_we \_should \_ch er ry - p ick \_. \\
\end{tabular}
\caption{Preprocessing of a parallel sentence pair using byte pair encoding (BPE).}
\label{tab:preprop}
\end{table*}

\begin{table}[t]
  \centering
  \begin{tabular}{ c  c  c  c c}
  Method & DeEn & EnFr & RoEn \\
  \hline
  \# Words Source & 50.5M & 20.0M & 11.7M \\
  \# Words Target & 53.2M & 23.6M & 11.8M \\
  \# Sentences  & 1.9M & 1.1M & 0.4M \\
\end{tabular}
\caption{Training data statistics.}
\label{tab:trainingdata}
\end{table}

\subsection{Test Data}
As test sets we use hand-aligned parallel sentences. For some of these data sets annotators were instructed to align target words that do not have a corresponding source word to a special null token. We always use the version without null tokens, i.e. a target word may not be aligned at all. The data sets are publicly available for German-English\footnote{\url{https://www-i6.informatik.rwth-aachen.de/goldAlignment/}}  and both for Romanian-English and English-French\footnote{\url{http://web.eecs.umich.edu/~mihalcea/wpt/index.html}}. Additionally, annotators made a distinction between probable and sure alignments. We evaluate the outputs of various systems based on the alignment error rate (AER) introduced by \citet{Och_2000}.

\section{Results}

\subsection{German-English}
We tune the hyperparameters of the approach presented in this paper on the German-English data set, which we choose as our development data (see Table \ref{tab:deen}). The na\"ive approach of averaging all the attention activations of the Transformer network yields sub-optimal results. For the unidirectional models it produces alignment error rates well above 50\%, combing two directions leads to 50.9\%. The additional alignment layer always improves results. While using the word embeddings (31.4\%) and the encoder output (28.6\%) as keys and values yields good improvements, using a combination of both works best (27.1\%). We speculate that the vectors representing the source tokens should both contain context and have a strong relation to the original word embedding. This procedure leads to results roughly as good as FastAlign (27.0\%). 

Optimizing the attention matrix from a random initialization to predict the reference translation with the SGD settings described in Section \ref{subsec:sgdInitialization} proves unsuccessful in terms of alignment
quality.
However, using the attention activations of the forward pass
noticeably improves the quality of the symmetrized alignments, yielding AER results similar to Giza++.

\begin{table}[t]
  \centering
  \begin{tabular}{ c  c  c  c }
  Method & DeEn & EnDe & Bidir \\
  \hline
  \hline
  Avg & 66.5\% & 57.0\% & 50.9\% \\
  Word & 36.9\% & 41.1\% & 31.4\% \\
  Enc & 39.2\% & 35.7\% & 28.6\% \\
  Add & 31.5\% & 34.7\% & 27.1\% \\
  \hline
  Rand+SGD & 65.9\% & 69.9\% & 61.3\% \\
  Add+SGD & 26.6\% & 30.4\% & 21.2\% \\
  \hline
  Giza++ & 21.0\% & 23.1\% & 21.4\% \\
  FastAlign & 28.4\% & 32.0\% & 27.0\% \\
  \hline
\end{tabular}
  \caption{AER results on the German-English task. The \emph{Bidir} column reports results 
           for the DeEn translation direction, obtained by symmetrizing both translation 
           directions with the grow-diag heuristic.}
\label{tab:deen}
\end{table}

\subsection{Qualitative Analysis}
\begin{figure*}[t!]
\centering
\begin{subfigure}{.24\textwidth}
  \centering
  \includegraphics[width=1.0\linewidth]{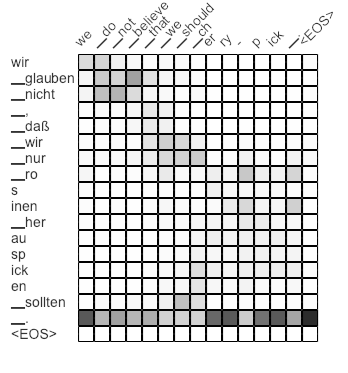}
  \caption{Avg (Soft)}
  \label{fig:sub1}
\end{subfigure}%
\begin{subfigure}{.24\textwidth}
  \centering
  \includegraphics[width=1.0\linewidth]{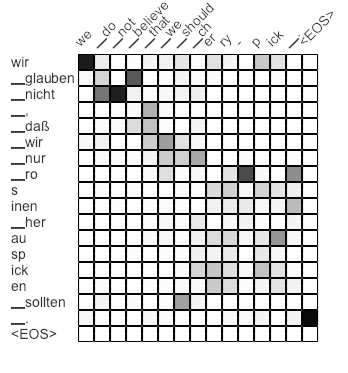}
  \caption{Add (Soft)}
  \label{fig:sub2}
\end{subfigure}%
\begin{subfigure}{.24\textwidth}
  \centering
  \includegraphics[width=1.0\linewidth]{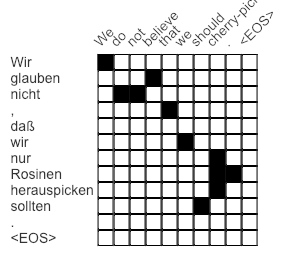}
  \caption{Add (Alignments)}
  \label{fig:sub4}
\end{subfigure}
\begin{subfigure}{.24\textwidth}
  \centering
  \includegraphics[width=1.0\linewidth]{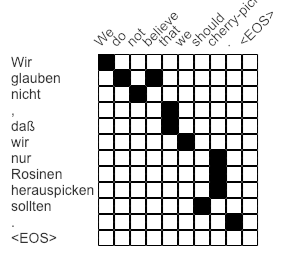}
  \caption{Reference}
  \label{fig:sub3}
\end{subfigure}%
\caption{%
\label{fig:softAttention}%
Visualization of average attention activations of different models when translating from German to English and the resulting alignments between words.}
\end{figure*}

We will analyse the resulting attention activations and alignments based on the example presented in Table \ref{tab:preprop}. This parallel sentence is the first sentence of our development set and highlights the following challenges: In both source (``wir'') and target (``we'') a word appears twice. Some words are not very common (``cherry-pick'') and get split into multiple subword units. The translation is non-literal, as ``Rosinen herauspicken'' (``pick raisins'') is translated as ``cherry-pick''.

We plot multiple average attention activations for this example in Figure \ref{fig:softAttention} with the a visualization tool provided by \citet{Rikters-EtAl2017PBML}. The Transformer mainly focuses its attention on the punctuation mark at the end of the sentence.\footnote{We never attend to the end of sentence symbol of the source sequence, because we mask the attention to it consistently during training and inference.} In contrast, the alignment layer attends to more meaningful source words. 

\begin{figure*}[t!]
\centering
\begin{subfigure}{.25\textwidth}
  \centering
  \includegraphics[width=1.0\linewidth]{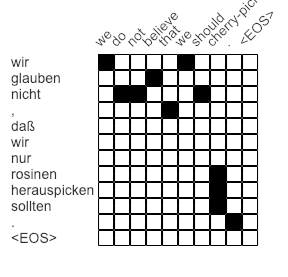}
  \caption{Rand+SGD 1}
\end{subfigure}%
\begin{subfigure}{.25\textwidth}
  \centering
  \includegraphics[width=1.0\linewidth]{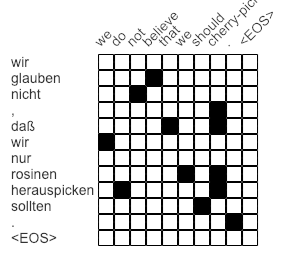}
  \caption{Rand+SGD 2}
\end{subfigure}%
\begin{subfigure}{.25\textwidth}
  \centering
  \includegraphics[width=1.0\linewidth]{pictures/deen-add-alignments.png}
  \caption{Add}
\end{subfigure}%
\begin{subfigure}{.25\textwidth}
  \centering
  \includegraphics[width=1.0\linewidth]{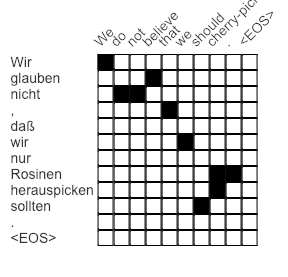}
  \caption{Add+SGD}
\end{subfigure}%
\caption{%
\label{fig:sgd}%
Comparison of word-based alignments when translating from German to English after optimizing the attention weights. We compare two different random initializations (Rand+SGD 1/2) and the intialization with the attention activations from the forward path (Add+SGD).}
\end{figure*}

Figure \ref{fig:sgd} shows alignments before and after applying SGD. When starting with a random initialization, no meaningful alignments can be generated. Note that different random initializations do not converge to similar alignments. However, when initializing with the attention activations of the forward pass, the resulting alignments improve in most of the cases. 

\subsection{English-French and Romanian-English}
We now test our approach on the English-French and Romanian-English test sets of \citet{Mihalcea_2003}. 
Similar to the German-English experiments, AER is consistently improved by adding the alignment layer and with optimization of the attention activations.

Interestingly, the neural approaches in this paper seem to profit more from symmetrizing both directions compared to the statistical approaches. The neural alignment models always use the full source context, but not the full target context, i.e. when generating an alignment we do not look at future target words. We speculate that this might be a contributing factor to the strong reduction in error rates by 
combining two unidirectional models.

We argue that the superior results of Giza++ on the English-French test set are mainly due to the large portion of probable alignment links (13{,}400 out of 17{,}438). This is favourable for Giza++, as it predicts the smallest number of alignments (20{,}200), while Add+SGD predicts considerably more (26{,}430). In contrast to the English-French test set, the Romanian-English set does not contain any probable alignments in its reference. 

\begin{table}[t]
  \centering
  \begin{tabular}{ c  c  c  c }
  Method & EnFr & FrEn & Bidir\\
  \hline
  \hline
  Avg & 55.4\% & 48.2\% & 31.4\% \\
  Add & 26.0\% & 26.1\% & 15.2\% \\
  Add + SGD & 23.8\% & 20.5\% & 10.0\% \\
  \hline
  Giza++ & 8.0\% & 9.8\% & 5.9\% \\
  FastAlign & 16.4\% & 15.9\% & 10.5\% \\
  \hline
\end{tabular}
\caption{AER results on the English-French task. The \emph{Bidir} column reports results 
         for the EnFr translation direction.}
\label{tab:enfr}
\end{table}

\begin{table}[t]
  \centering
  \begin{tabular}{ c  c  c  c }
  Method & RoEn & EnRo & Bidir\\
  \hline
  \hline
  Avg & 45.7\% & 52.2\% & 39.8\% \\
  Add & 37.7\% & 38.7\% & 32.8\% \\
  Add + SGD & 32.3\% & 34.8\% & 27.6\% \\
  \hline
  Giza++ & 28.7\% & 32.2\% & 27.9\%\\
  FastAlign & 33.8\% & 35.5\% & 32.1\%\\
  \hline
\end{tabular}
\caption{AER results on the Romanian-English task. The \emph{Bidir} column reports results 
           for the RoEn translation direction.}
\label{tab:roen}
\end{table}

\section{Conclusion}
This paper addresses the problem of extracting meaningful word alignments from the self-attentive
Transformer neural machine translation model. We extend the network with an alignment layer 
that contains no skip connections around the encoder-attention sub-layer and thus is encouraged to 
learn to attend to source words that correspond to the current target word.
We further introduce a novel inference procedure to query the model with a given target word.
By symmetrizing alignments extracted from models for both translation directions we achieve
an alignment quality that is comparable to IBM Model 4 as implemented in Giza++ on two of the three
tasks. Different from previous work our model is trained in an unsupervised fashion and does
not require injecting external knowledge from the IBM models into the training pipeline.

\bibliographystyle{acl_natbib}
\bibliography{naaclhlt2019}

\appendix


\end{document}